\documentclass[lettersize,journal]{IEEEtran}
\usepackage{amsmath,amsfonts}
\usepackage{algorithmic}
\usepackage{algorithm}
\usepackage{array}
\usepackage[caption=false,font=normalsize,labelfont=sf,textfont=sf]{subfig}
\usepackage{textcomp}
\usepackage{stfloats}
\usepackage{url}
\usepackage{verbatim}
\usepackage{graphicx}
\usepackage{cite}
\usepackage{paralist}
\usepackage{hyperref}
\usepackage{color}

\usepackage{xspace}

\makeatletter
\DeclareRobustCommand\onedot{\futurelet\@let@token\@onedot}
\def\@onedot{\ifx\@let@token.\else.\null\fi\xspace}

\def\eg{\emph{e.g}\onedot} 
\def\ie{\emph{i.e}\onedot} 
 
\def\etc{\emph{etc}\onedot}

\makeatother

\begin{document}

\title{Image-based Geolocalization by Ground-to-2.5D Map Matching}

\author{Mengjie Zhou, Liu Liu, Yiran Zhong, Andrew Calway
        % <-this % stops a space
\thanks{Mengjie Zhou and Andrew Calway are with the School of Computer Science, University of Bristol, Bristol, United Kingdom.}
\thanks{Liu Liu is with Huawei Cyberverse Dept., Beijing, China.}
\thanks{Yiran Zhong is with Shanghai AI Lab, Shanghai, China.}
\thanks{Corresponding author: Yiran Zhong (zhongyiran@gmail.com).}
%\thanks{Manuscript received xxx, xxxx; revised x, 2021.}}
}

\markboth{Journal of \LaTeX\ Class Files,~Vol.~14, No.~8, August~2021}%
{Shell \MakeLowercase{\textit{et al.}}: A Sample Article Using IEEEtran.cls for IEEE Journals}


\maketitle

\begin{abstract}
We study the image-based geolocalization problem, aiming to localize ground-view query images on cartographic maps. Current methods often utilize cross-view localization techniques to match ground-view query images with 2D maps. However, the performance of these methods is unsatisfactory due to significant cross-view appearance differences. In this paper, we lift cross-view matching to a 2.5D space, where heights of structures (\eg, trees and buildings)  provide geometric information to guide the cross-view matching. We propose a new approach to learning representative embeddings from multi-modal data. 
Specifically, we establish a projection relationship between 2.5D space and 2D aerial-view space. The projection is further used to combine multi-modal features from the 2.5D and 2D maps using an effective pixel-to-point fusion method.
By encoding crucial geometric cues, our method learns discriminative location embeddings for matching panoramic images and maps.
Additionally, we construct the first large-scale ground-to-2.5D map geolocalization dataset to validate our method and facilitate future research. 
Both single-image based and route based localization experiments are conducted to test our method. Extensive experiments demonstrate that the proposed method achieves significantly higher localization accuracy and faster convergence than previous 2D map-based approaches.
\end{abstract}


\section{Introduction}
\label{sec:intro}
\begin{figure}[t]
  \centering 
  \includegraphics[width=1.0\linewidth]{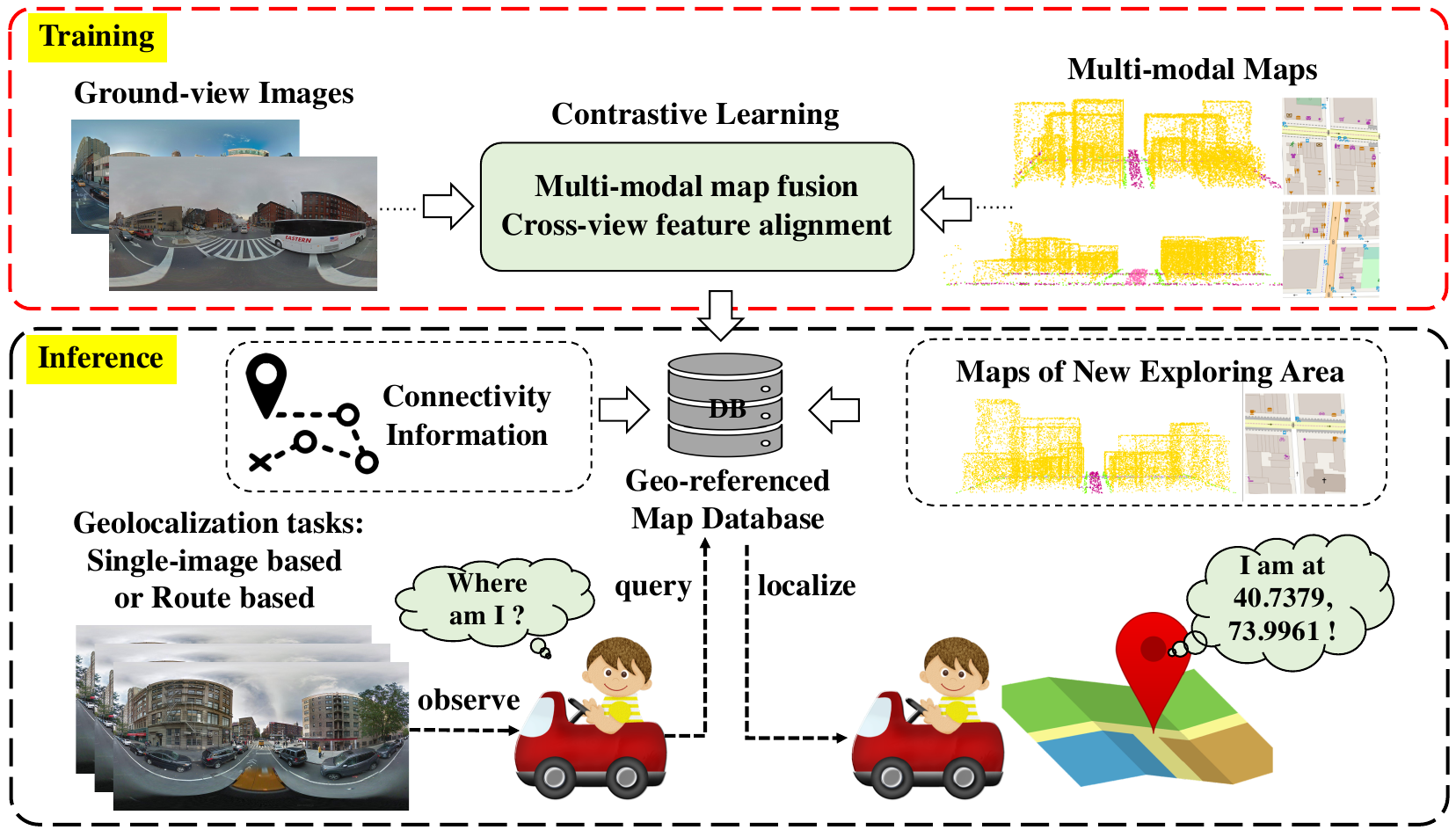}
  \vspace{-6mm}
  \caption{{Illustration of query ground-view image and multi-modal map for the geolocalization task. During the training phase, the precollected ground-view images and processed multi-modal maps are utilized as the input for the contrastive learning architecture to achieve multi-modal fusion and cross-view feature alignment. The well-trained model and map data, including connectivity information from an unknown environment, are then used to establish a geo-referenced database for the online image-based geolocalization task. The semantic category uniquely encodes the color of the point cloud, as shown in Fig.~\ref{fig:semantic}.}}
  \vspace{-4mm}
  \label{fig:intro}
\end{figure}

\IEEEPARstart{W}{e} study the problem of image-based geolocalization using ground-to-2.5D map matching. Given a ground-view query image, we aim to estimate the geospatial position where the query image is taken. This is done by querying the ground-view image with respect to a large-scale and georeferenced multi-modal map database consisting of 2.5D structural map models and 2D aerial-view map tiles. An example scenario of such a ground-to-2.5D map cross-view localization problem is illustrated in Fig.~\ref{fig:intro}.

Most state-of-the-art cross-view localization methods~\cite{vo2016localizing,hu2018cvm,shi2019spatial,shi2020looking,shi2020optimal,liu2019lending,sun2019geocapsnet,NEURIPS2021_f31b2046} employ a map with satellite/aerial RGB images for retrieval. Though effective, they have two principal limitations: i) The appearance of satellite map images changes with seasonal (summer, winter, \etc) and illumination (day, night, \etc) conditions. Furthermore, it also differs across satellites and covers dynamic objects such as cars and trees, bringing challenges for robust long-term localization; and ii) significant cross-view appearance differences. Since satellite view captures an image orthogonal to the ground plane, only the highest landmarks along the vertical direction are observable, whereas ground view can see the side views of these landmarks. The above-mentioned significant viewpoint difference presents significant challenges for matching cross-view images.

This paper addresses the two limitations mentioned above by using georeferenced multi-modal maps. We propose using 2D cartographic maps instead of satellite/aerial RGB images because they are both robust to radiance and time changes, and are compact. In addition, we include the height information of structures into the map to bridge the domain gap between cross-view images, yielding the 2.5D map models. Compared with detailed 3D models, the 2.5D model is compact and easy to achieve while still containing enough structure information for cross-view matching. It is worth noting that 2.5D map models are now enabled by the majority of mapping service providers, such as Google Maps and OpenStreetMap, and can be obtained easily.

Having both 2D maps and 2.5D maps, how to learn discriminative feature embeddings for each multi-modal map to enable ground-view image retrieval? In this paper, we propose to fuse 2D maps and 2.5D maps in the same feature space with effective fusion techniques. Specifically, we first design a data processing pipeline to automatically extract 2.5D map models from OpenStreetMap and convert them to point clouds using the surface sampling strategy. We then  build a triplet-like architecture with InfoNCE loss~\cite{oord2018representation} to learn an embedding space for intra- and inter-modal discrimination.  

Ground-to-2.5D map geolocalization is a non-trivial task, \ie, the significant difference in the appearance of panoramas and maps and the feature fusion between the image domain and the 2.5D map domain. 
We incorporate geometric clues through explicit geometric transformations—polar transforms—and implicit geometric feature learning of 2.5D maps in order to close the cross-view gap. Our results show that the use of 2.5D maps leads to improved performance.
We examine various fusion methods for multi-modal fusion and discover that pixel-to-point feature fusion delivers superior performance.
To evaluate our method and facilitate the research, we constructed the first large-scale ground-to-2.5D map geolocalization dataset, which consists of 113767 panoramic images and geo-tagged maps from the cities of New York and Pittsburgh. There are three testing sets split for evaluation, each containing 5000 locations, covering trajectories of 69.3 $km$ to 75.6 $km$. We perform two types of localization: single-image based localization and route based localization, using the extracted location embeddings to validate our method. Extensive experiments show that our multi-modal map based localization methods achieve higher localization accuracy than state-of-the-art methods~\cite{samano2020you}. 
In summary, the main contributions of this work are:
\begin{compactitem}
    \item A 2.5D map based cross-view matching method, enabling accurate long-term cross-view localization;
    \item A multi-modal feature extraction method, fusing features from 2D maps and 2.5D maps;
    \item A large-scale 2.5D map based cross-view localization dataset, consisting of 113767 panoramic images and geo-tagged multi-modal maps, covering multiple cities;
    \item State-of-the-art localization accuracy with two 2.5D map based cross-view localization: single-image based and route based localization, demonstrating the effectiveness of using 2.5D maps for cross-view localization.
\end{compactitem}

\begin{figure*}[t]
  \centering \includegraphics[width=0.9\linewidth]{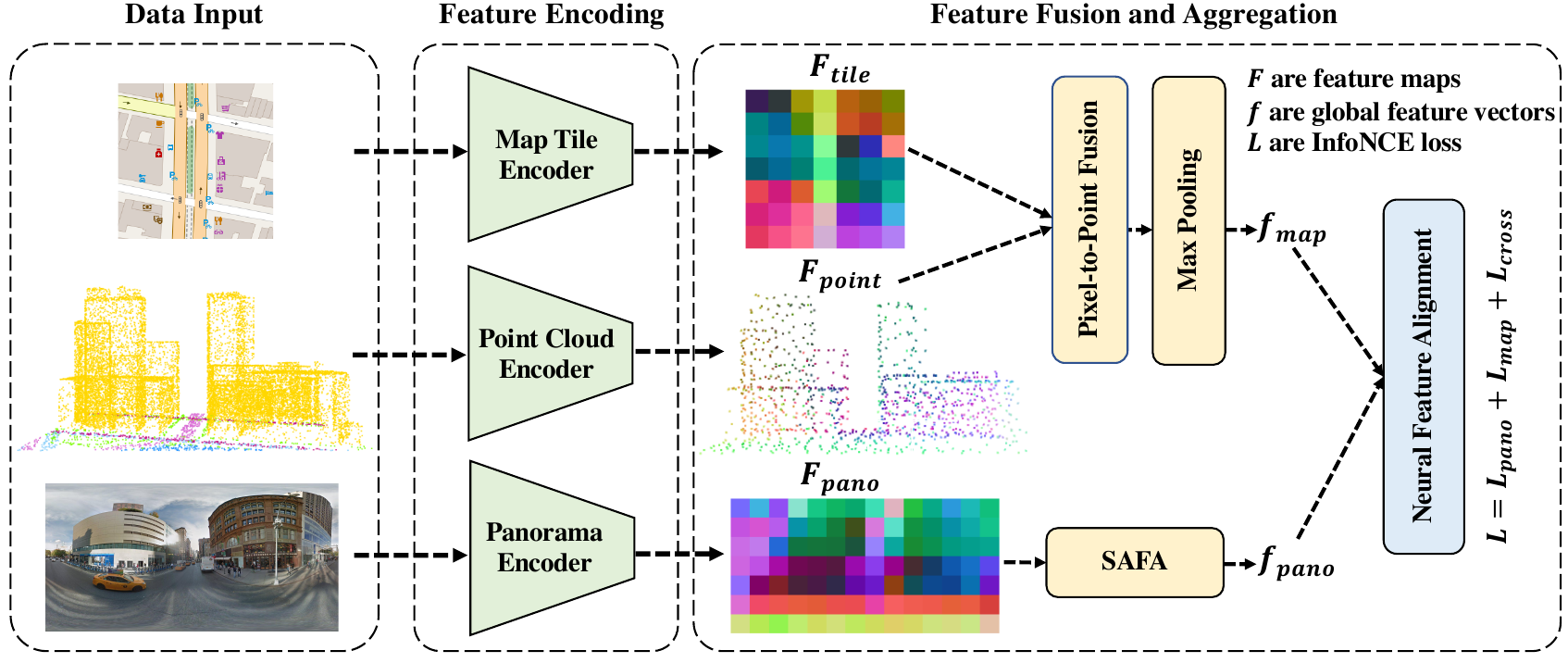}
  \vspace{-3mm}
  \caption{{The overall network architecture consists of a map tile branch, a point cloud branch, and a panorama branch. Each branch consists of an independent feature encoder. The fusion block employs the pixel-to-point projection from 2D space to 2.5D space. The feature aggregators, max pooling, and spatial-aware feature aggregation (SAFA) produce the global feature vectors, which embed semantic and geometric information to achieve neural feature alignment via contrastive learning. The color of the input point cloud is uniquely encoded by the semantic category as shown in Fig.~\ref{fig:semantic}. Each feature map is projected to the RGB space via principal component analysis (PCA) for visualization. } 
  }
  \label{fig:model}
\end{figure*}

\section{Related Work}
\label{sec:literature}
Image-based geolocalization has been extensively studied for years. There have been a significant number of papers published on this topic and we only cite some of the works that we consider most related to our method. We also briefly review some point-cloud processing methods for the sake of completeness.

\vspace{2mm}
\noindent
\textbf{Cross-view Geolocalization}
To tackle the data availability problem, using dense satellite imagery as the reference database has become an attractive geolocalization approach. The main challenge is feature extraction and similarity matching across views. Due to drastic appearance and viewpoint differences, traditional hand-crafted features obtain unsatisfactory performance~\cite{bansal2011geo,lin2013cross}. With the booming of deep learning, researchers begin to explore effective deep neural networks and efficient learning strategies for cross-view geolocalization. 
Efforts are mainly taken to develop task-related network layers ~\cite{hu2018cvm,shi2019spatial,shi2020optimal,shi2020looking,NEURIPS2021_f31b2046}, effective triplet loss~\cite{hu2018cvm,sun2019geocapsnet}, large datasets~\cite{liu2019lending}, and geometric transformation to bridge the cross-view gap~\cite{liu2019lending,shi2019spatial}.

\vspace{2mm}
\noindent
\textbf{Map-related Task}
Publicly available map data, such as OpenStreetMap (OSM), has been used for self-driving vehicles~\cite{brubaker2015map,ma2017find,SeffX16}. 
Inspired by the cross-view works, Panphattarasap and Calway~\cite{panphattarasap2018automated} first proposed to use 2D OSM maps as the reference database for the geolocalization task. To achieve high scalability, an extremely compact 4-bit descriptor indicating the presence or not of semantic features (junctions and building gaps) is designed to represent locations. 
Then, Samano et al.~\cite{samano2020you, zhou2021efficient} generalized the approach in~\cite{panphattarasap2018automated} by linking images to 2D OSM maps in an embedding space. 
% our idea, 2.5D
Not limited to the usage of 2D maps, researchers are also exploring the benefits of higher dimensional maps. 
Given an initial coarse GPS signal, Anil et al.~\cite{armagan2017learning} and Hai et al.~\cite{hai2021bdloc} achieved global localization using 2.5D building maps.
Although GPS is a low-cost localization signal, it frequently loses accuracy in challenging environments such as urban canyons because it is susceptible to atmospheric uncertainty, building blockage, multi-path bounced signals, and signal interference. 
While \cite{armagan2017learning,hai2021bdloc} use GPS as a prior and refines the results with images, our work aims to use image-based geolocalization techniques to do initial positioning, particularly in situations where on-device GPS signal is unavailable.

\vspace{2mm}
\noindent
\textbf{Point Cloud Representation Learning} 
The 2.5D map data is an untextured 3D map
constructed from a 2D cadastral map with heights, which can be typically represented in the form of point cloud, mesh and voxel. 
Compared with mesh and voxel, the point cloud is friendly for network processing, generalization, efficient storage, and broad usage.
% pointnet
To directly process this irregular geometric data structure with neural networks, ~\cite{qi2017pointnet} first proposed a unified architecture named PointNet which is robust to the permutation variance of the point cloud input.
% pointnet++
However, PointNet treats each point independently and doesn't explore the local neighborhood information. Therefore, ~\cite{qi2017pointnet++} proposed a hierarchical neural network named PointNet++ which applies PointNet~\cite{qi2017pointnet} recursively on multiple point cloud subsets partitioned by metric space distance. 
% DGCNN
Similarly, the method named DGCNN~\cite{wang2019dynamic} also proposed to incorporate local neighboring information to enrich the representation power. The difference is that the DGCNN establishes the topological link of the neighborhood in feature space, rather than the metric space used in the PointNet++~\cite{qi2017pointnet++}. It is indicated that the feature space can capture semantic characteristics over potentially long distances. 
% POINT TRANSFORMER
In recent years, self-attention networks have revolutionized natural language processing and image analysis. Motivated by this impressive development, ~\cite{zhao2021point} proposed the Point Transformer which introduces self-attention layers for point cloud representation learning.
It is indicated that the self-attention operator is especially suitable for point cloud processing because of its permutation and cardinality invariance to the input elements. 
These methods have shown their effectiveness in the subsequent tasks of 3D shape classification and scene segmentation. Moreover, our research extends their applicability to localization tasks, further confirming their robustness and versatility across various applications.

\section{Network Architecture}
\label{sec:method}
\subsection{Overall network architecture}
\label{sec:network}
% no weight sharing, similarity and difference
The overall network architecture for learning location embeddings is illustrated in Fig.~\ref{fig:model}. It is structured in a triplet-like shape with three individual branches, namely, Map Tile Branch, Point Cloud Branch, and Panorama Branch. The two upper branches are used to learn multi-modal map features and the bottom branch is used to learn semantic features from panoramic images. 
All learned features from various modalities are then utilized for subsequent neural feature alignment by employing contrastive learning in an embedding space.
It is worth noting that there is no weight sharing between branches because each one processes information that is vastly different from the others. In the following sections, we provide technical details of each branch and our feature fusion strategies.

\subsection{Map tile branch}
The map tile branch is mainly used to extract features from the map tile input, which is an image of a local region of the 2D map.
The map tile encoder is built upon the ResNet18 network~\cite{he2016deep}, including four convolutional blocks to produce a 512-channel feature volume $\mathbf{F}_\text{tile}$ with a resolution that is 1/32 of the original input. 

% point cloud branch (encoder + projection)
\subsection{Point cloud branch}
The 2.5D map we use is an untextured map constructed from a 2D cadastral map augmented with height information, which significantly reduces storage memory requirements and transmission bandwidth compared with fully textured 3D models. The 2.5D structural map model can be processed in a variety of ways. We use the point cloud form due to its simplicity and conducive to network processing.

We process the 2.5D map in the point cloud branch. The feature encoder is built upon popular backbones used for point cloud representation learning. In this paper, we study both MLP-based~\cite{qi2017pointnet,qi2017pointnet++,wang2019dynamic} and MLP-Transformer~\cite{zhao2021point} based structure as the feature encode backbones and demonstrate the consistency of performance improvement brought by the 2.5D map. 
After the point cloud encoder, the original input is encoded into a shape of $N \times C_{3D}$ feature volume $\mathbf{F}_\text{point}$, where $N$ is the number of points and $C_{3D}$ is the number of channels. 

\subsection{Multi-modal fusion}
\label{subsec:fusion}
The output features from the map tile branch and point cloud branch 
are fused for the subsequent multi-modality feature learning. 
We study different fusion strategies, \ie, global fusion, point-to-pixel fusion and pixel-to-point fusion, and find that \textbf{pixel-to-point fusion} brings the best performance. A detailed comparison is provided in the experiment section.

\begin{figure*}[t]
  \centering \includegraphics[width=0.95\linewidth]{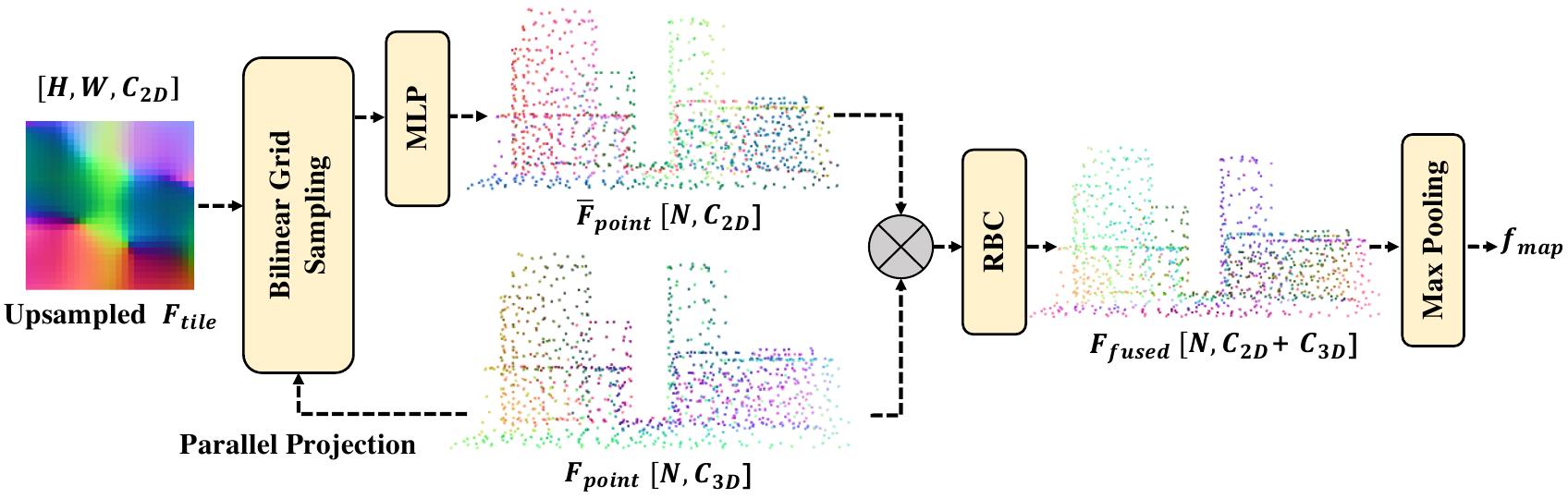}
  \vspace{-3mm}
  \caption{Pixel-to-Point Fusion. To create a global semantic feature vector, we use bilinear grid sampling and parallel projection to project upsampled tile features $\mathbf{F}_{\text{tile}}$ to the same shape as point cloud features $\mathbf{F}_{\text{point}}$. We then concatenate and fuse these features using Conv$1\times1$-BN-ReLU operations (RBC) and a max pooling aggregator. To visualize each feature map, we project it into the RGB space using principal component analysis (PCA).}
  \label{fig:2to3_fusion}
\end{figure*}
The low spatial resolution of the feature map, as indicated in~\cite{liu2021learning}, has an impact on point-to-pixel or pixel-to-point knowledge transfer. To recover the spatial resolution, we perform an additional bilinear upsampling operation after the map tile feature encoder to quadruple the size of the feature map. Furthermore, we incorporate an additional projection module that begins and ends with fully connected layers and includes batch normalization (BN), ReLU activation, and dropout layer~\cite{srivastava2014dropout} in the middle. This operation reduces the feature dimension of the point cloud to that of a map tile.

Before entering the fusion block, the feature volume $\mathbf{F}_\text{tile}$ ($H \times W \times C_{2D}$) and $\mathbf{F}_\text{point}$ ($N\times C_{3D}$) have been obtained through individual encoders, upsampling and projection modules. To achieve the pixel or point level fusion, we establish a parallel projection relationship between 2D aerial-view space and 2.5D space:
\begin{gather}
\label{eq:projection}
        x_i = (\overline{x}_i+0.5W_g-C_x)\frac{(W-1)}{(W_g-1)} \\
        y_i = (\overline{y}_i+0.5H_g-C_y)\frac{(H-1)}{(H_g-1)}
\end{gather}
where $(\overline{x}_i,\overline{y}_i)$ is the point coordinate, and $(x_i,y_i)$ is the projected pixel coordinate. ($W$,$H$) and ($W_g$,$H_g$) are the size of the feature map in pixel and geographic level, respectively, while $(C_x, C_y)$ represents the geographical coordinate of central point.

Subsequently, as depicted in Fig.~\ref{fig:2to3_fusion}, we generate the projected feature volume $\mathbf{\overline{F}}_\text{point}$ ($N\times C_{2D}$) through bilinear grid sampling at $(\overline{x},\overline{y})$ with the feature volume $\mathbf{F}_\text{tile}$. This projected volume is then concatenated with $\mathbf{F}_\text{point}$ after passing through a multi-layer perceptron (MLP) including three Conv$1\times1$-BN-ReLU blocks.
Finally, an additional Conv$1\times1$-BN-ReLU block and a max pooling operator are applied to fuse and aggregate multi-modal feature volume, processing it into a unified global feature vector with the desired embedding size. As highlighted in~\cite{qi2017pointnet}, max pooling, being a symmetric function, is well-suited for processing unordered point cloud data.

\subsection{Panorama branch}
\begin{figure}[t]
  \centering \includegraphics[width=1.0\linewidth]{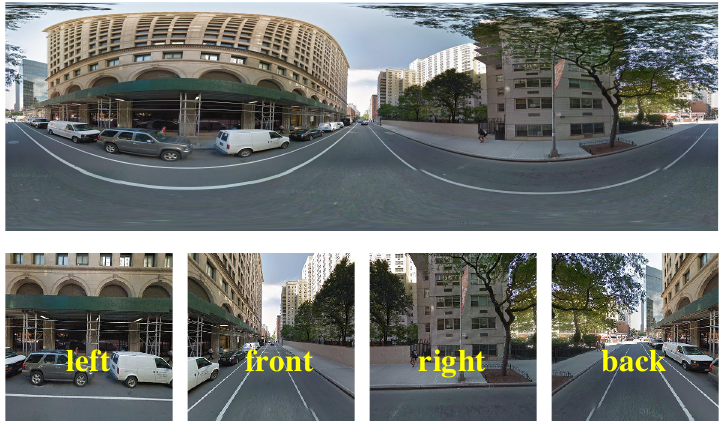}
  \caption{Illustration of a panoramic image (upper) and its four cropped snapshots (bottom) facing to the front, back, left and right.}
  \label{fig:panorama}
\end{figure}

Given the specific heading angle, previous works~\cite{panphattarasap2018automated,samano2020you,zhou2021efficient} choose to crop the ground-view panoramic images into four orthogonal views using Equirectangular Projection as shown in Fig.~\ref{fig:panorama}. 
Although the angle of view as seen by a human is preserved in this method, the structural information of the scene, such as the height of the buildings, is incomplete. However, this is not the case for 2.5D maps. Consequently, to avoid the risk of information loss and potential mismatch between query panorama and referenced map data, we choose to feed the original panorama directly into the panorama branch.

We use ResNet50 as the panorama encoder as suggested in~\cite{samano2020you}.
After passing through four convolutional blocks, the input panoramic image is transformed into a 512-channel feature volume with a 1/32 resolution of the original size. 
% SAFA
We leverage the spatial-aware feature aggregation (SAFA) module~\cite{shi2019spatial} to localize the salient features and encode the relative spatial layout information.

\begin{figure*}[t]
  \centering \includegraphics[width=1.0\linewidth]{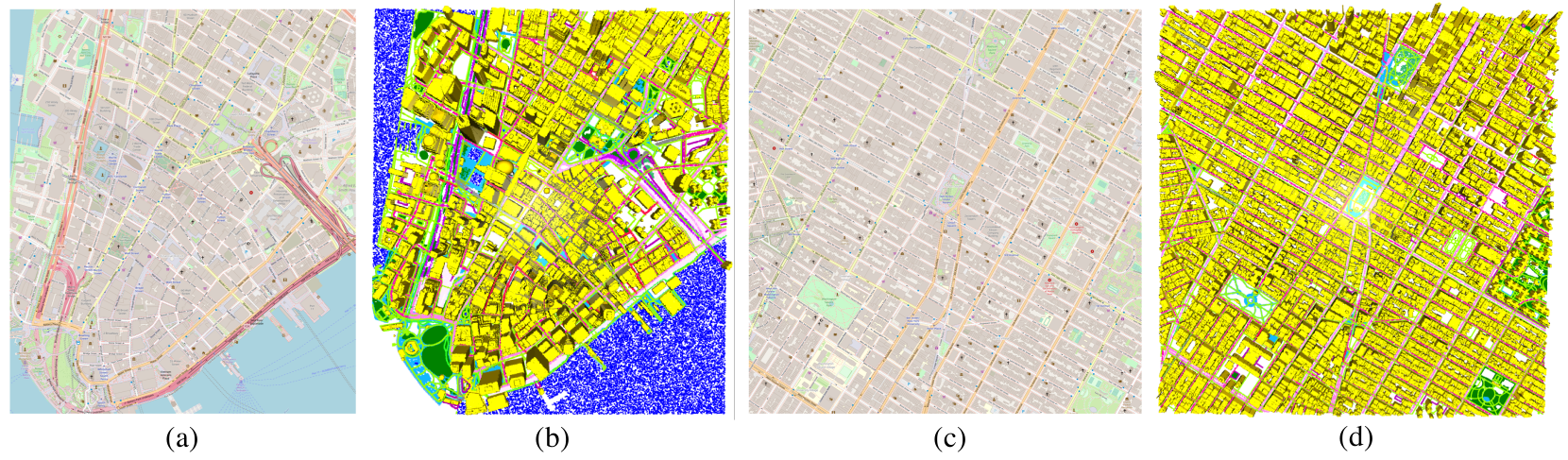}
  \caption{Multi-modal map dataset. The 2D map (a) and 2.5D map (b) are from the area of Wall Street, which includes narrow streets and highways with irregular intersections. The 2D map (c) and 2.5D map (d) are from the area of Union Square, which includes densely distributed skyscrapers, brownstones, townhouses, and parks located on regular street grids. 
  For the 2.5D map, the unique color is encoded by the semantic category as shown in Fig.~\ref{fig:semantic}.}
  \label{fig:dataset}
\end{figure*}

\section{Model Training}
\label{sec:training}
% augmentation
Our model is trained in an end-to-end way via contrastive learning. We combine intra-modal and inter-modal discrimination to formulate the loss function during training, which is inspired by the pioneering work~\cite{afham2022crosspoint}. 
As demonstrated in~\cite{huang2021towards}, richer data augmentation implies better generalization for contrastive self-supervised learning. 
Given an input panoramic image $\mathbf{I}_i$, we construct augmented versions $\mathbf{I}_i^{t_1}$ and $\mathbf{I}_i^{t_2}$ using transformations such as rotation, color jittering, normalization, erasing and Gaussian noising in sequence. Similarly, the augmented versions $\mathbf{M}_i^{t_1}$ and $\mathbf{M}_i^{t_2}$ of the map tile $\mathbf{M}_i$ are constructed using transformations such as normalization, erasing and Gaussian noising. For the point cloud $\mathbf{P}_i$,  $\mathbf{P}_i^{t_1}$ and $\mathbf{P}_i^{t_2}$ are constructed using random shuffle, jittering, and points removing in a sequential manner. All corresponding transformation parameters are generated randomly using uniform distribution in small ranges to ensure positive alignment. 

% intra
After the encoding and aggregation module, the global feature vectors of $\mathbf{I}_i^{t_1}$ and $\mathbf{I}_i^{t_2}$ are extracted which we denote as $\mathbf{q}_i^{t_1}$ and $\mathbf{q}_i^{t_2}$. By using the fusion block, we get the fused global feature vector $\mathbf{r}_i^{t_1}$ for $(\mathbf{M}_i^{t_1},\mathbf{P}_i^{t_1})$ and $\mathbf{r}_i^{t_2}$ for $(\mathbf{M}_i^{t_2},\mathbf{P}_i^{t_2})$. The optimization goal is to maximize the similarity of positive pairs while minimizing the similarity of negative pairs in a mini-batch. 
For the panorama-modal discrimination, the loss is calculated as:
\begin{equation}
    \mathcal{L}_{\text{pano}} = \frac{1}{2B}\sum\limits_{i=1}^{B}\left[l(\mathbf{q}_i^{t_1},\mathbf{q}_i^{t_2})+l(\mathbf{q}_i^{t_2},\mathbf{q}_i^{t_1})\right]
\end{equation}
The loss of the map-modal discrimination is calculated as:
\begin{equation}
    \mathcal{L}_{\text{map}} = \frac{1}{2B}\sum\limits_{i=1}^{B}\left[l(\mathbf{r}_i^{t_1},\mathbf{r}_i^{t_2})+l(\mathbf{r}_i^{t_2},\mathbf{r}_i^{t_1})\right]
\end{equation}
The loss of the cross-modal discrimination is calculated as:
\begin{gather}
    \mathcal{L}_{\text{cross}} = \frac{1}{2B}\sum\limits_{i=1}^{B}\left[l(\mathbf{q}_i,\mathbf{r}_i)+l(\mathbf{r}_i,\mathbf{q}_i)\right] \\
    \mathbf{q}_i = \frac{1}{2}(\mathbf{q}_i^{t_1}+\mathbf{q}_i^{t_2}) \\
    \mathbf{r}_i = \frac{1}{2}(\mathbf{r}_i^{t_1}+\mathbf{r}_i^{t_2})
\end{gather}
We leverage the InfoNCE loss~\cite{oord2018representation} as the function of $l(\mathbf{z}_i,\mathbf{h}_i)$ for the positive pair of $\mathbf{z}_i$ and $\mathbf{h}_i$:
\begin{equation}\label{eq::infonce}
    l(\mathbf{z}_i,\mathbf{h}_i) = -\log\frac{\exp(d(\mathbf{z}_i,\mathbf{h}_i)/\tau)}{\sum\limits_{k=1}^{B}\exp(d(\mathbf{z}_i,\mathbf{h}_k)/\tau)}
\end{equation}
where $B$ is the mini-batch size, $\tau$ is the temperature co-efficient, and $d(.)$ is the cosine similarity function, which executes the dot product between $L_2$ normalized feature vector. Finally, the overall loss function is formulated as:
\begin{equation}
    \mathcal{L} = \mathcal{L}_\text{pano} + \lambda_1\mathcal{L}_\text{map} + \lambda_2\mathcal{L}_\text{cross}
\end{equation}
where $\lambda_1$ and $\lambda_2$ are weighting factors to control the influence of each loss component, which we set to be equal as suggested in~\cite{afham2022crosspoint,samano2020you}.

\begin{figure*}[t]
  \centering \includegraphics[width=0.9\linewidth]{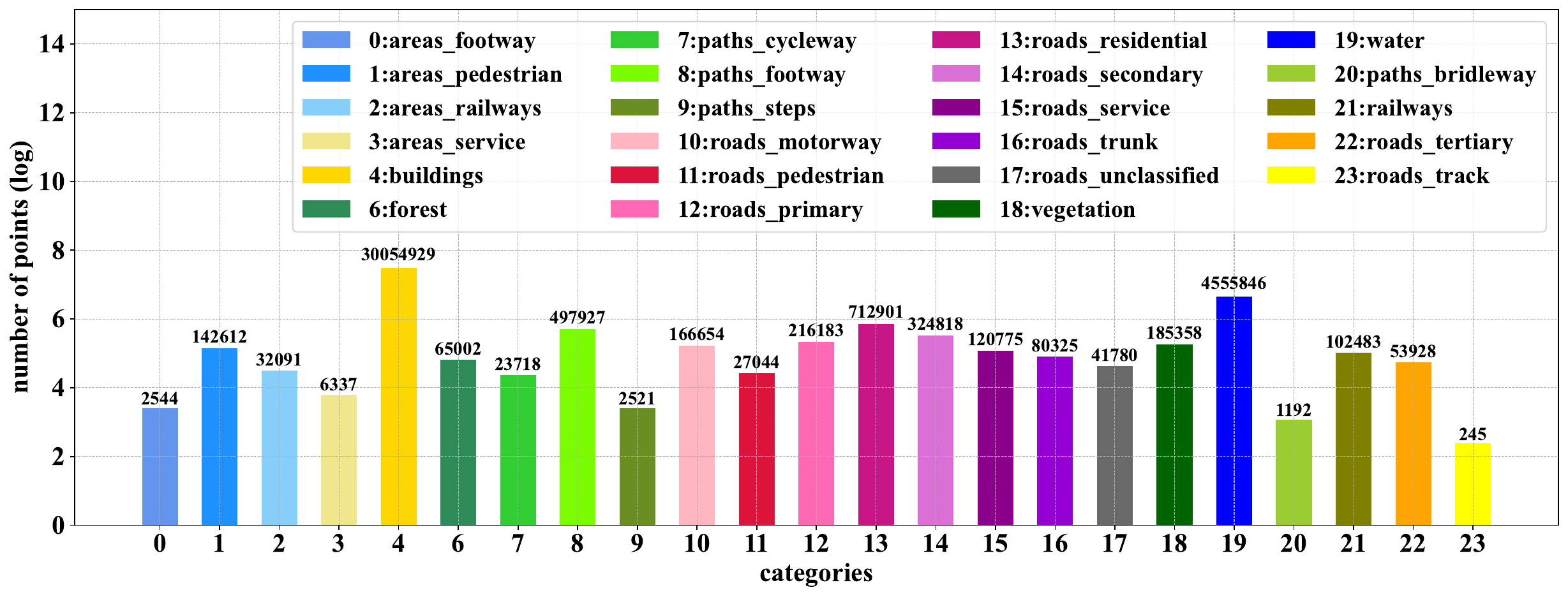}
  \caption{A statistical overview for the number of points distribution across different semantic categories within the geographical area of Manhattan. Semantic categories are initially labeled 0 to 23 and then encoded as 24-D one-hot vectors for network input.
  Note that Category 5 (Coastline) is absent in Manhattan's semantic categories.
  For ease of display, we project the data into log space but annotated the actual number of points at the top of each bar.
}
  \label{fig:semantic}
\end{figure*}

\section{The Dataset}
\label{sec:datasets}
% StreetLearn, training, testing split
To evaluate our method and facilitate the research, we construct a large-scale ground-to-2.5D map geolocalization dataset. 
The ground-view images are collected from the StreetLearn dataset~\cite{streetlearnwebpage,mirowski2019streetlearn}, consisting of 113767 panoramic images named with unique string identifiers in the cities of New York (Manhattan) and Pittsburgh. In the metadata, there is detailed information about the geographical position (lat/long coordinates and altitude in meters), camera orientation (pitch, roll, and yaw angles), and the connected neighbors of each location. 
To generate the training/testing/validation split, we use the same approach proposed in~\cite{samano2020you}. 
There are two testing sets from areas of Union Square and Wall Street, each containing 5000 locations, covering around 75.6 $km$ and 73.1 $km$ trajectories, respectively. The validation set is generated from the area of Hudson River with the same size as the testing set, covering around 69.3 $km$ trajectory. 
There are diverse scenes in different areas, including skyscrapers, highways, parks, and riversides located on regular street grids (Union Square, Hudson River) or narrow streets with irregular intersections (Wall Street).

The multi-modal map data is automatically generated from the public map service, OpenStreetMap~\cite{OSMwebpage}, as illustrated in Fig.~\ref{fig:dataset}.
The 2D map tiles with the size of $256\times256$ are rendered using Mapnik~\cite{Mapnikwebpage}. 
Specifically, the center of each map tile corresponds to the geo-tagged location, and the upward direction of each map tile is aligned with the vehicle heading direction. 
We design a data processing pipeline to automatically process the 2.5D structure map model from the OpenStreetMap (OSM) to the point cloud. 
Specifically, we first render the OSM metadata of each semantic category into a triangle-mesh structural model using Blender~\cite{blender-osmwebpage}, then uniformly sample points on triangles using the Barycentric coordinate system~\cite{meyer2002generalized}. 
The number of points to be sampled is determined by the sampling density (0.1 in this paper) and surface area. Defining the vertices of a triangle surface as $\mathbf{v}_1, \mathbf{v}_2,\mathbf{v}_3\in R^3$, the area $A$ and the number of sampled points $N$ are calculated as:
\begin{gather}
    A = \frac{1}{2}\Vert (\mathbf{v}_1-\mathbf{v}_3)\times(\mathbf{v}_2-\mathbf{v}_3) \Vert_2 \\
    N = \text{density}\cdot{A},
\end{gather}
and $N$ new points $\mathbf{p}_i$ are sampled as:
\begin{equation}
    \mathbf{p}_i = (1-\sqrt{r_1^i})\mathbf{v}_1+\sqrt{r_1^i}(1-r_2^i)\mathbf{v}_2+\sqrt{r_1^i}r_2\mathbf{v}_3
\end{equation}
where $r_1$ and $r_2$ are two random variables uniformly distributed from 0 to 1. 
Finally, we merge points of each semantic category to a completed point cloud covering the whole area and crop the corresponding 2.5D map of a small region given the geo-tagged location.  
In this work, the multi-modal map data represent a local area with the geographical size of $152\times152$ $m^2$.
Totally, there are 98767 ground-view image and multi-modal map pairs for training, 5000 pairs for validation, and 10000 pairs for testing. 
The dataset and code are available at \url{https://github.com/ZhouMengjie/2-5DMap-Dataset}.

\section{Experiments}
\label{sec:exp}
\subsection{Setting}
We implement our network in Pytorch~\cite{paszke2019pytorch}. All models are trained in an end-to-end manner for 60 epochs on 4 Nvidia A100 GPUs. We empirically select ImageNet~\cite{deng2009imagenet} pre-trained weights to initialize the map tile encoder and Places365~\cite{zhou2017places} for the panorama encoder. The spatial-aware feature aggregation module is initialized with a normal distribution. All other parameters are initialized using a uniform distribution.

Before entering the network, the panorama and map tile is resized to $448\times224$ and $224\times224$ respectively. The dense point cloud is firstly normalized to the range of -1 to 1 and then downsampled to 1024 points with the farthest point sampling strategy as suggested in~\cite{qi2017pointnet,qi2017pointnet++,wang2019dynamic,zhao2021point}. 
{The network output initially has an embedding size of 4096. To minimize redundancy and enhance computation and storage efficiency, we use the Principal Component Analysis (PCA) method for flexible feature dimension reduction. This results in a final embedding size of either 128 or 16, depending on the localization methods used.

During back-propagation, we use the Adaptive Sharpness-aware Minimization (ASAM) strategy combined with the AdamW optimizer to optimize the network. The AdamW optimizer has an initial learning rate of $1\times10^{-4}$ and a weight decay of $0.03$. It has been shown that using the ASAM strategy leads to a significant improvement in the model's generalization performance.}
The cosine annealing scheduler~\cite{loshchilov2016sgdr} is used to gradually decrease the learning rate to a minimum (0 in this paper). We use a batch size of $32$ and the temperature in Eq.~\eqref{eq::infonce} is set to $0.07$. The model performing best on the validation set is chosen for the localization tasks.

\subsection{Geolocalization results}
\label{sec:single}
We validate our learned location embeddings in two localization strategies, \ie, single-image based localization and route based localization. For the former, we use the Top-$k$ recall rate to evaluate the geolocalization performance on the Hudson River, Wall Street, and Union Square. That is, given a query panoramic image, we retrieve the Top-$k$ geo-tagged reference maps by measuring the similarity ($L_2$ distance) between their 128-D (Dimension) global semantic features. If the matching reference map is ranked within the Top-$k$  list, a query panoramic image is considered to be localized successfully. The Top-$k$ recall rate shows the percentage of correctly localized queries. 
For the latter, we use 500 randomly generated routes consisting of 40 adjacent locations in the area of Hudson River, Wall Street and Union Square. The test data is provided by work~\cite{samano2020you}, and the distance between each location is around 10 meters. We adopt the Top-1 recall rate as our evaluation metric, which is measured by the percentage of correctly localized routes as a function of route length.
Specifically, a route is considered to have been successfully localized at step $t$ if and only if the matching reference maps from step $t-4$ to $t$ are all ranked first. 

\vspace{2mm}
\noindent
\textbf{Single-image based localization}
\begin{figure*}[t]
    \centering  
    \hspace{-5mm}
    \subfloat[]{\includegraphics[width=0.32\textwidth]{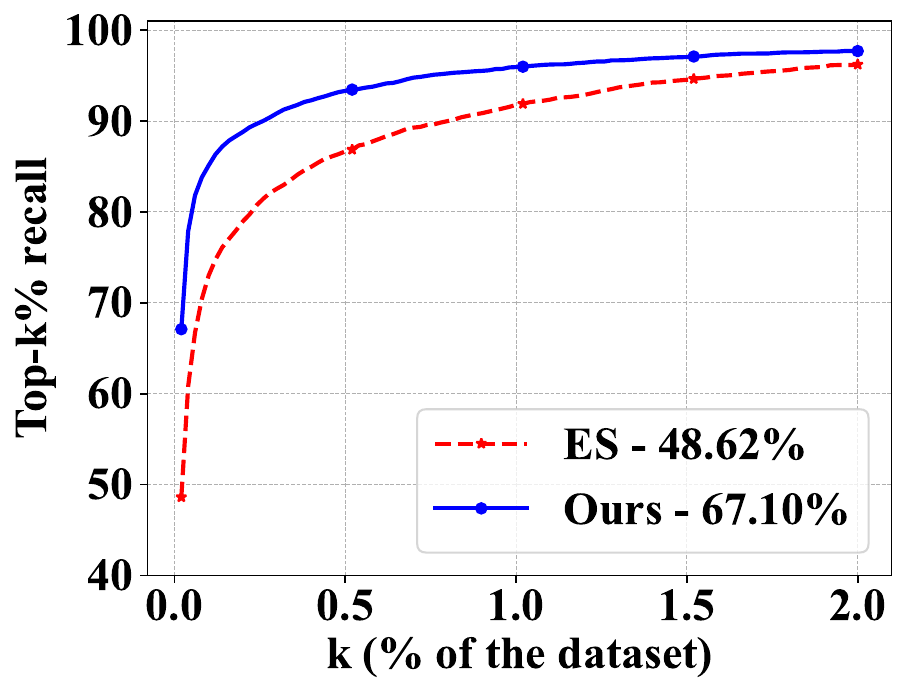}}
    \subfloat[]{\includegraphics[width=0.32\textwidth]{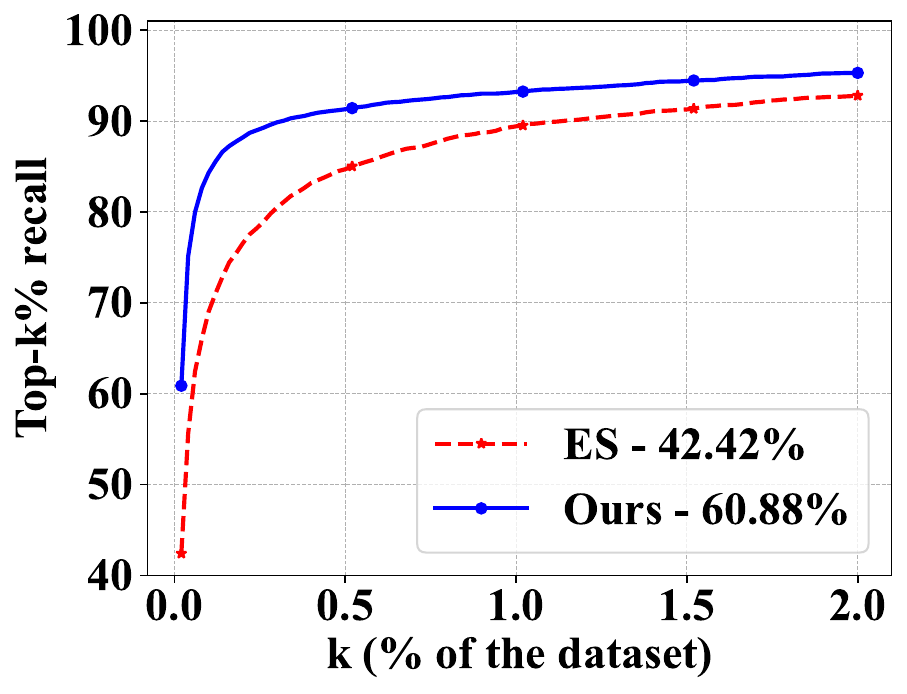}}
    \subfloat[]{\includegraphics[width=0.32\textwidth]{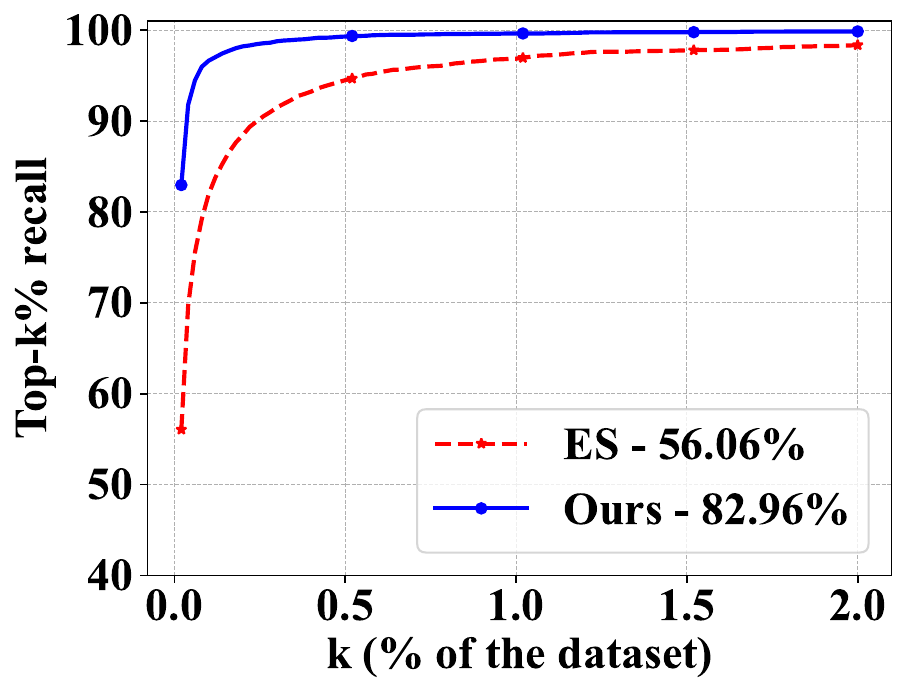}}
    \caption{Comparison between single-modal method and multi-modal method on the single-image based localization task. We use Embedding Space Descriptor (ES)~\cite{samano2020you} as the single-modal reference method. The Top-$k$\% recall rate is calculated to evaluate the localization performance in the area of Hudson River (a), Wall Street (b), and Union Square (c). The Top-1 recall rate is presented in the lower-right legend.}
    \label{fig:single_localization}   
\end{figure*}
{In our study, we evaluated the recall rate for the top $k$\% of the dataset, where $k$\% represents a fraction of the dataset size. To establish a baseline, we included the state-of-the-art single-modal method~\cite{samano2020you}. For our multi-modal fusion strategy, we utilized pixel-to-point fusion. Our results indicate that using 2.5D maps can yield significantly better performance compared to single-modal methods. Specifically, using the DGCNN~\cite{wang2019dynamic} as the point cloud encoder resulted in the greatest performance gains, with improvements of 19.08\% for Hudson River, 18.24\% for Wall Street, and 26.9\% for Union Square. Fig.~\ref{fig:single_localization} presents our quantitative results for single-image based localization.}
 
\begin{figure*}[t]
  \centering \includegraphics[width=0.9\linewidth]{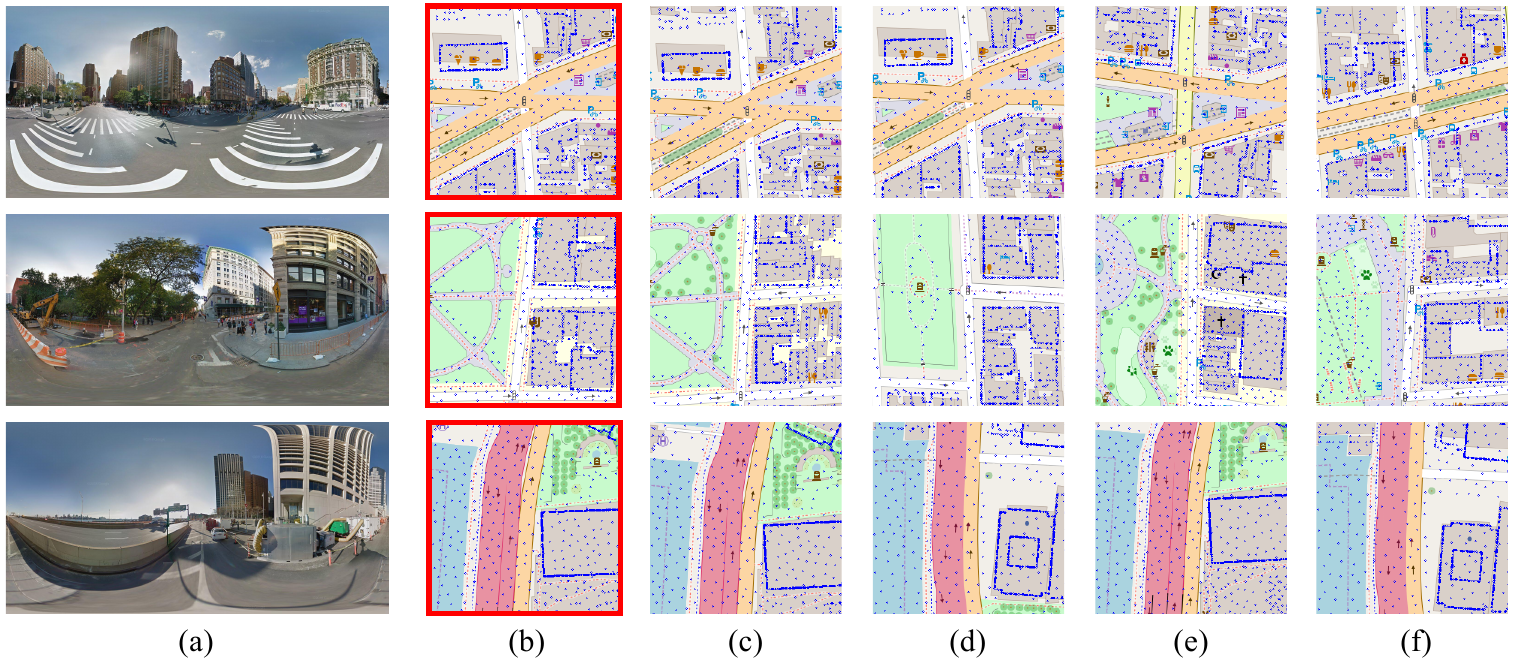}
  \caption{Top-5 retrieved maps (b)-(f) given a query panoramic image (a). The correct related map of the query is outlined in red.}
  \label{fig:retrieving}
\end{figure*}

We also illustrate examples of query panoramic images and the Top-5 retrieved maps in Fig.~\ref{fig:retrieving}. The corresponding map of each location image is outlined in red. The successful localization in these challenging environments indicates that our model has learned representative semantic features from both the panorama and map domains. 

\vspace{2mm}
\noindent
\textbf{Route based localization}
\begin{figure*}[t]
    \centering  
    \hspace{-5mm}
    \subfloat[]{\includegraphics[width=0.33\textwidth]{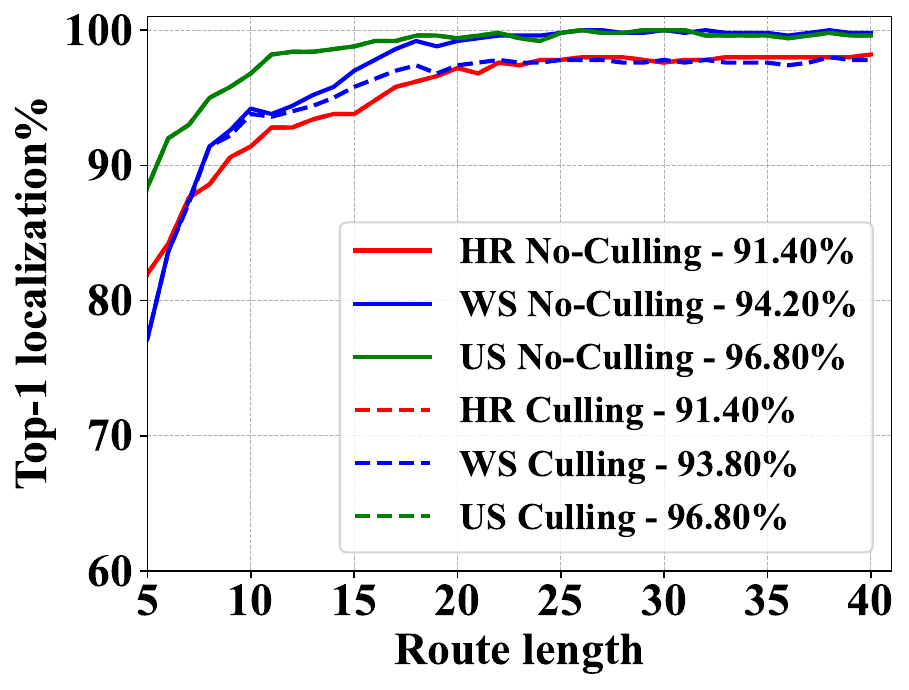}}
    \subfloat[]{\includegraphics[width=0.33\textwidth]{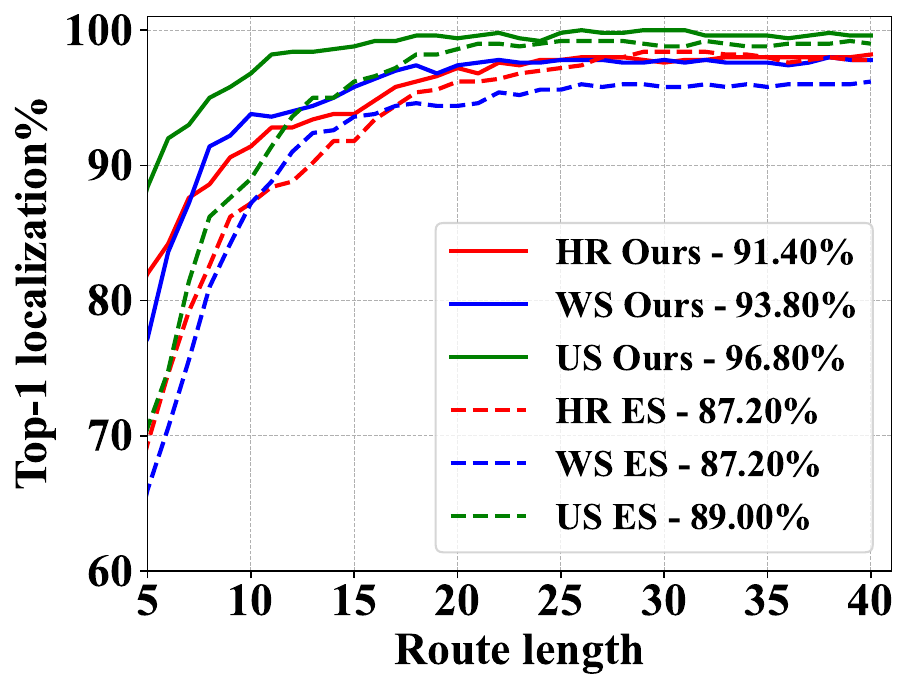}}
    \subfloat[]{\includegraphics[width=0.33\textwidth]{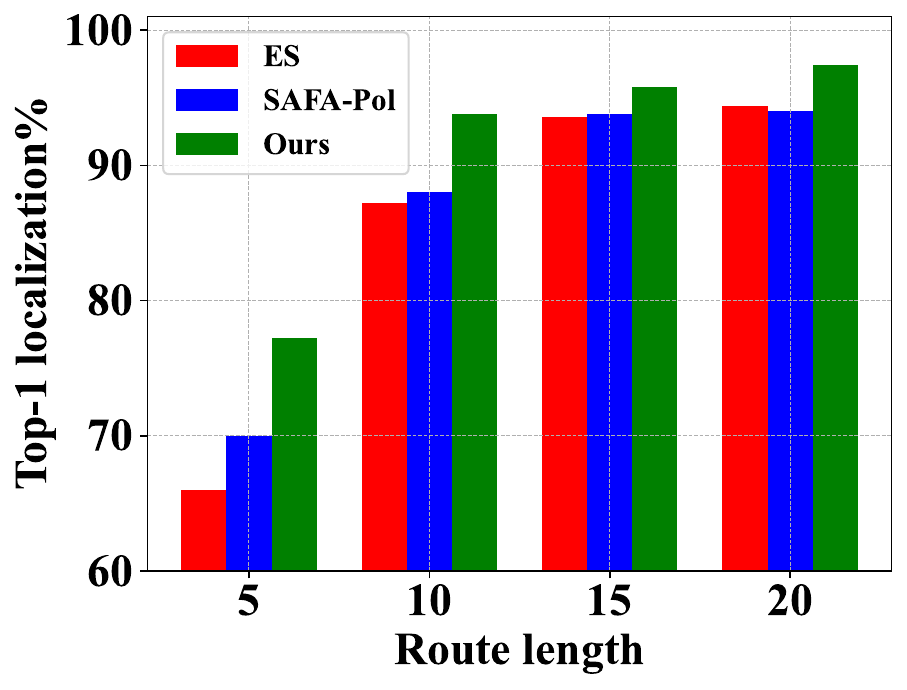}}
    \caption{The performance of route based localization. (a) shows the comparison with and without the culling strategy. (b) shows the comparison between single-modal and multi-modal methods in three different areas. (c) shows the comparison between various map-based methods on Wall Street. HR, WS, and US separately represent the Hudson River, Wall Street, and Union Square. The Top-1 accuracy at step 10 is shown in the lower-right legend.}
    \label{fig:localization}   
\end{figure*}
Route based localization is often used to localize in large areas, as a single descriptor is not sufficiently discriminative in large cities with a variety of repeated scene settings. We implement a route based localization method that is proposed in~\cite{panphattarasap2018automated} with efficient modifications, \ie, rather than storing all route candidates in advance, we generate candidate routes online based on connectivity information between adjacent locations. To further improve the algorithm's performance, we adopt a culling strategy to ensure localization efficiency.

Fig.~\ref{fig:localization} (b) shows the performance of our method in route based localization. Compared with the state-of-the-art~\cite{samano2020you} in a single modal, our method achieves notably better performance. When moving to the location with a route length of 5, the multi-modal method already achieves over 75\% localization accuracy, which is more than 10\% higher than the single-modal method. The results indicate that the fusing of multi-modal map features for the route based localization task can achieve higher accuracy and faster convergence speed. 

\subsection{Ablation study}
\begin{figure*}[t]
    \centering  
    \hspace{-5mm}
    \subfloat[]{\includegraphics[width=0.33\textwidth]{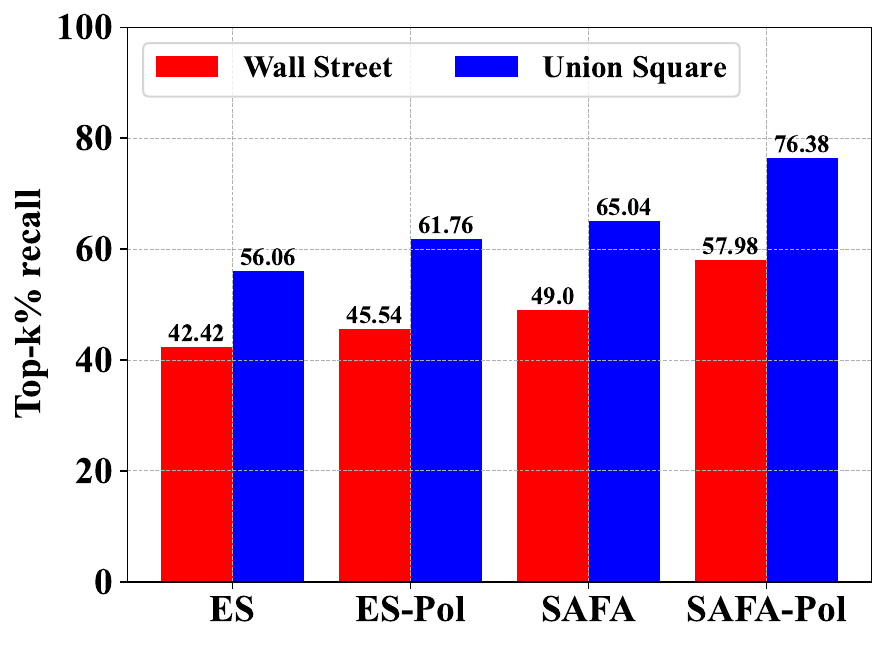}}
    \subfloat[]{\includegraphics[width=0.33\textwidth]{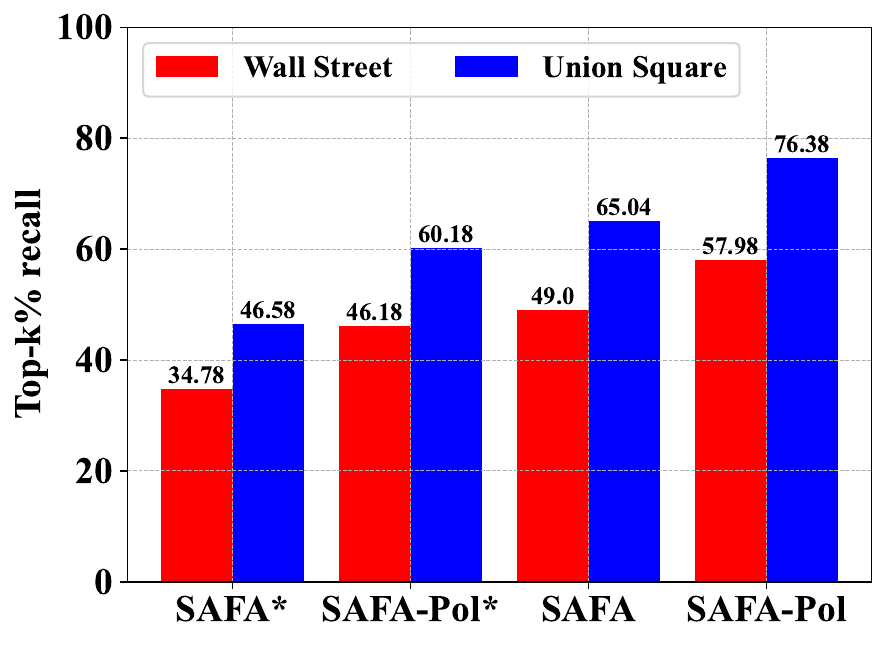}}
    \subfloat[]{\includegraphics[width=0.33\textwidth]{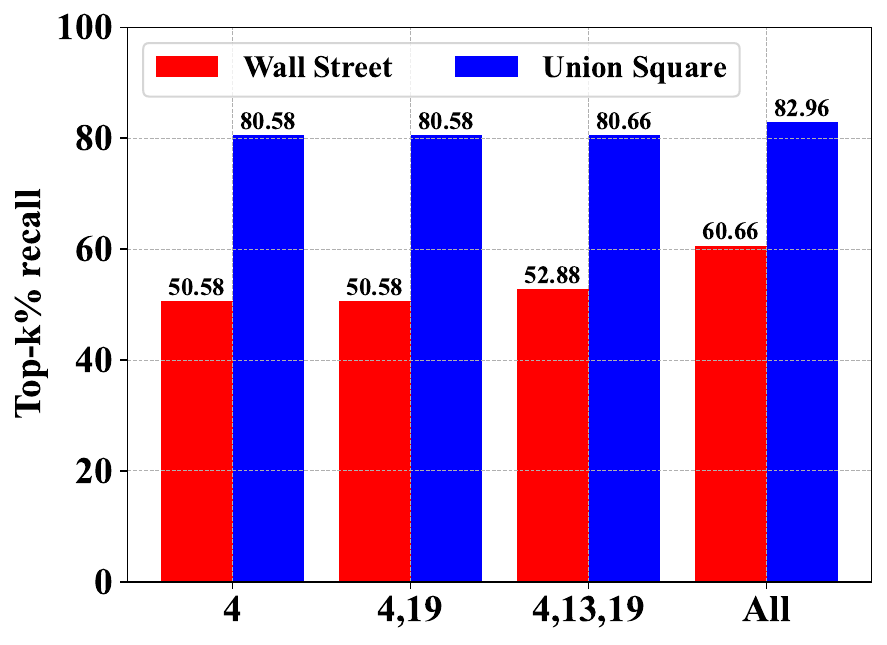}}
    \caption{The performance of single-image based localization. (a) shows the comparison with and without SAFA/Polar transform (Pol denotes polar transform). (b) shows the comparison with different optimizers (* denotes using Adam optimizer). (c) shows the comparison with different 2.5D map inputs, including specific semantic categories (4, 13, 19 referring to buildings, residential roads and water respectively as shown in Fig.~\ref{fig:semantic}).}
    \label{fig:agg_opt}   
\end{figure*}

\noindent
\textbf{Feature aggregation and polar transform}
{We replace the flatten operation in the prior work of \cite{samano2020you} with the spatial-aware feature aggregation (SAFA) technique. As shown in Figure~\ref{fig:agg_opt}(a), SAFA delivers remarkable performance improvements of 6.58\% on the Wall Street dataset and 8.98\% on the Union Square dataset.}

% SAFA and polar
{To mitigate the cross-view discrepancy between ground-view and aerial-view images, previous methods often use polar transform to coarsely align the geometric configuration between the two views~\cite{shi2019spatial,shi2020looking,toker2021coming}. Since both map tiles and satellite images share the same viewing angle, we apply the same explicit geometric transformation on map tiles for single-modal method, and use it as an improved version of baseline to do a comparison with our proposed multi-modal method achieved by implicit geometric relationship learning.}

{We have observed that the polar transformation has advantageous effects on both single-modal methods, with or without SAFA. When combined with SAFA, the polar transform provides even higher performance gains of 8.98\% and 11.34\% on two different testing areas. These results confirm the effectiveness of SAFA in mitigating the impacts of features distorted by the polar transform proposed in~\cite{shi2019spatial}. 
Similar trends are present in route-based localization, as demonstrated in Fig.~\ref{fig:localization} (c). These findings have encouraged us to include SAFA in our multi-modal methods as well.}

\vspace{2mm}
\noindent
\textbf{Optimization}
% adam, asam 
{we investigate the performance disparity resulting from utilizing different optimization strategies, specifically Adam and Adaptive Sharpness-Aware Minimization (ASAM). Fig.~\ref{fig:agg_opt}(b) presents compelling evidence of a substantial performance improvement, with a remarkable 14.22\% and 18.46\% gain observed for SAFA, and 11.8\% and 16.2\% for SAFA-Pol. These results emphasize the necessity of devising more suitable and effective training strategies to achieve significantly enhanced localization performance.
Analogous outcomes are also evident in route-based localization when compared with experimental results reported in the original map-based study~\cite{samano2020you}.}

\begin{figure}[t]
  \centering 
  \vspace{-3mm}
  \includegraphics[width=1.0\linewidth]{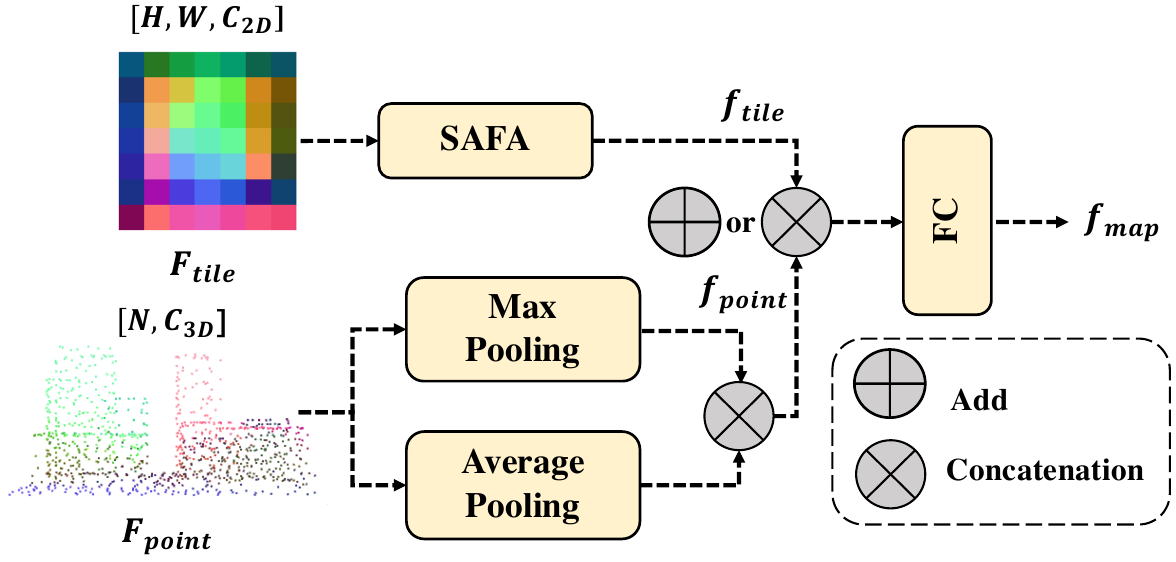}
  \caption{Procedure of global fusion. The feature volume outputs from separate feature encoders are initially aggregated into single-modal global feature vectors. Subsequently, these individual global feature vectors are combined through either concatenation or addition, resulting in a fused global feature vector after passing through a fully-connected layer.}
  \vspace{-4mm}
  \label{fig:global_fusion}
\end{figure}

\vspace{2mm}
\noindent
\textbf{Fusion strategy}
To study the fusion of multi-modal features, we examine three design options: global fusion with add or concatenation operators, point-to-pixel fusion, and pixel-to-point fusion. The global fusion block is illustrated in Fig.~\ref{fig:global_fusion}. Initially, a map tile encoder extracts the feature volume $\mathbf{F}_\text{tile}$, which is then fed into a spatial-aware feature aggregation (SAFA) strategy to create a $C_{g}$-channel global feature vector $\mathbf{f}_\text{tile}$. Similarly, the point cloud encoder extracts the feature volume $\mathbf{F}_\text{point}$, which is then projected into two $C_{g}/2$-channel global vectors using max and average pooling operations. These vectors are concatenated to form a $C_{g}$-channel global feature vector $\mathbf{f}_\text{point}$. Finally, the multi-modal global feature vectors are either concatenated or added along the channel dimension and projected to the desired embedding size after a fully connected layer.

\begin{figure}[t]
  \centering \includegraphics[width=1.0\linewidth]{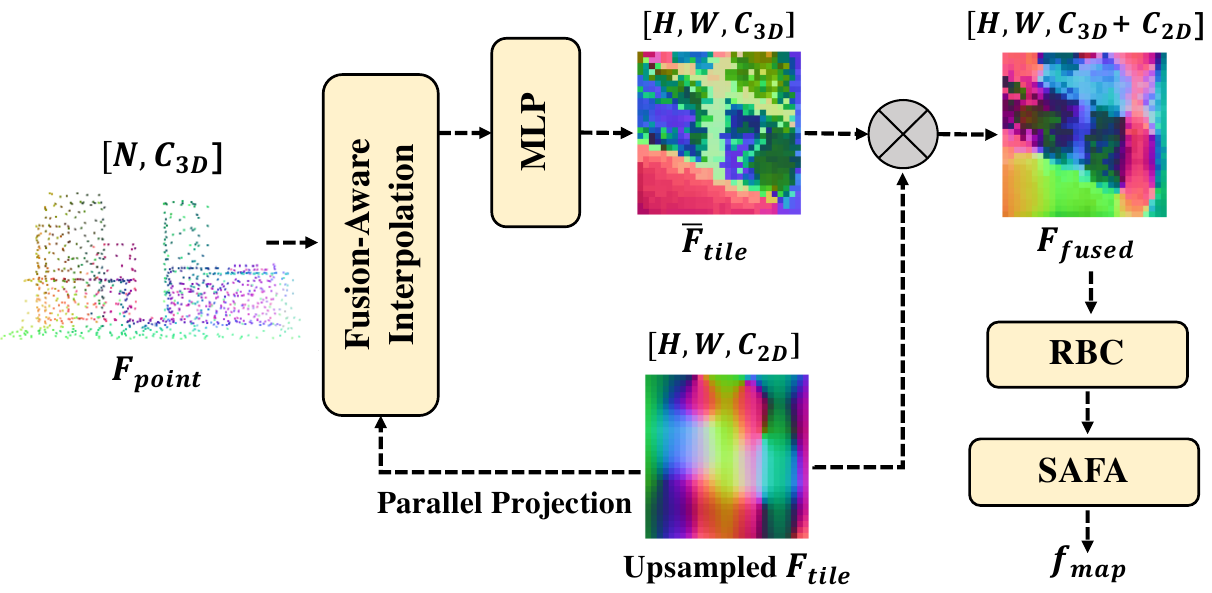}
  \caption{Point-to-Pixel Fusion. Using the fusion-aware interpolation~\cite{liu2022camliflow} and parallel projection, the point cloud features $\mathbf{F}_{\text{point}}$ are projected into the same shape of map tile features $\mathbf{F}_{\text{tile}}$, then concatenated and fused with map tile features $\mathbf{F}_{\text{tile}}$ to generate a global semantic feature vector $\mathbf{f}_{\text{map}}$ using the sequential Conv$1\times1$-BN-ReLU operations (RBC) and spatial-aware feature aggregation.}
  \label{fig:3to2_fusion}
\end{figure}

\begin{table}
\caption{Comparison between multi-modal methods using global fusion with concatenation and add operator, point-to-pixel fusion and pixel-to-point fusion in different testing areas. Method denoted with * utilize the four-fold upsampled map tile feature as an input to the fusion block.}
\begin{center}
\small
\setlength{\tabcolsep}{2mm}
\begin{tabular}{lccc}
\hline
Fusion Strategy & Hudson River & Wall Street & Union Square \\
\hline
Concatenate  & 63.38 & 56.98 & 74.10 \\
Add  & 64.08 & 56.26 & 76.54 \\
Point-to-Pixel  & 64.82 & 57.32 & 76.58 \\
Pixel-to-Point & 66.96 & 60.00 & 81.50 \\
Point-to-Pixel*  & 65.80 & 58.20 & 79.64 \\
\textbf{Pixel-to-Point*} & \textbf{67.70} & \textbf{60.66} & \textbf{82.96} \\
\hline
\end{tabular}
\end{center}
\label{table:fusion}
\end{table}

The point-to-pixel fusion block is shown in Fig.~\ref{fig:3to2_fusion}. 
Similar to the pixel-to-point fusion method, we fed the extracted feature volumes $\mathbf{F}_\text{tile}$ ($H \times W \times C_{2D}$) and $\mathbf{F}_\text{point}$ ($N \times C_{3D}$) into the fusion block, along with the parallel projection relationship between 2D aerial-view space and 2.5D space.
In the fusion block, an interpolated feature volume $\mathbf{\overline{F}}_\text{tile}$($H\times W\times C_{2D}$) is first generated by fusion-aware interpolation~\cite{liu2022camliflow} at $(x,y)$ with the feature volume $\mathbf{F}_\text{point}$, and concatenated with $\mathbf{F}_\text{tile}$ after a multi-layer perceptron (MLP), consisting of three Conv$1\times1$-BN-ReLU blocks. Next, after a Conv$1\times1$-BN-ReLU block and spatial-aware feature aggregation module, the fused feature volume is projected into a unified global feature vector with the desired embedding size.

Table.~\ref{table:fusion} illustrates the Top-1 recall rate localizing in Hudson River, Wall Street, and Union Square. 
As shown, the pixel-to-point fusion with upsampled map tile features exhibits the highest success rate across all testing areas. For instance, when compared to global fusion using the add operator, there are notable performance gains of 3.62\%, 4.4\%, and 6.42\% observed in different localization areas.
Furthermore, in comparison to the explicit geometric transformation method SAFA-Pol, the pixel-to-point fusion strategy has a 2.86\%, 2.68\%, and 6.58\% higher success rates in separate testing areas for Top-1 accuracy.
Given the effectiveness of the pixel-to-point fusion strategy, it has been selected as our primary feature fusion approach, unless stated otherwise.

\begin{figure}[t]
  \centering \includegraphics[width=1\linewidth]{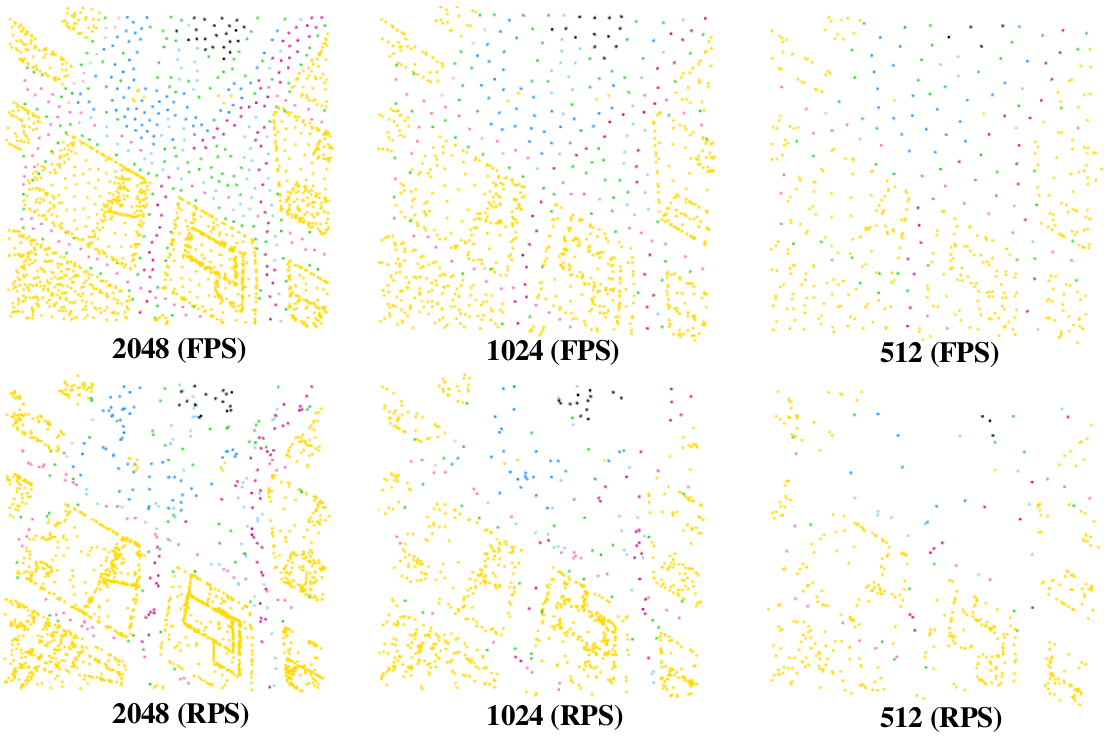}
  \caption{Aerial-viewed point cloud data generated by farthest point sampling (FPS) and random point sampling (RPS). The color is uniquely encoded by the semantic category as shown in Fig.~\ref{fig:semantic}.}
  \label{fig:sampling}
\end{figure}

\vspace{2mm}
\noindent
\textbf{Point sampling strategy}
We conduct a comparison between two point cloud sampling strategies – farthest point sampling (FPS) and random point sampling (RPS). We sampled 256, 512, 1024, and 2048 points for each strategy to process the point cloud. Based on the data in Table~\ref{table:robustness}, we find that FPS generally provides better localization accuracy, and increasing the number of sampled points results in better performance. This is likely because FPS preserves more structure information compared to RPS. We include a visualization of the differences in Fig.~\ref{fig:sampling}. After considering the trade-off between efficiency and accuracy, we decided to use FPS with 1024 points as our sampling strategy.

\begin{table}
\caption{Robustness of multi-modal method to density variation and the number of points. Various types of point clouds are generated by the farthest point sampling and random point sampling in the area of Union Square. The Top-1 recall rate (\%) is calculated to evaluate the localization performance.}
\begin{center}
\small
\begin{tabular}{lcccc}
\hline
Sampling Strategy & 256 & 512 & 1024 & 2048 \\
\hline
\textbf{Farthest Point Sampling}  & \textbf{73.58} & \textbf{81.12} & \textbf{82.96} & \textbf{83.66} \\
Random Point Sampling  & 49.72 & 67.50 & 77.10 & 80.70 \\
\hline
\end{tabular}
\end{center}
\label{table:robustness}
\end{table}

\vspace{2mm}
\noindent
\textbf{Point cloud encoder}
{We study multi-modal methods utilizing different point cloud encode backbones. Specifically, we implement pixel-to-point fusion for Pointnet~\cite{qi2017pointnet} and DGCNN~\cite{wang2019dynamic} based methods since they do not employ any further point sampling during the forward pass. For Pointnet++\cite{qi2017pointnet++} and Point Transformer\cite{zhao2021point}, the number of points is reduced to 256 and 16, respectively. It is important to be aware that this particular process has been known to cause a notable decrease in performance, based on previous experiments. Therefore, we choose to utilize global fusion with the addition operator for these two methods in order to evaluate their performance.
Table~\ref{table:backbone} presents the outcomes of our evaluations. When combined with global fusion, employing Point Transformer as the point cloud encoder yields the best performance. When adopting pixel-to-point fusion, using DGCNN as the point cloud encoder achieves superior results.
In conclusion, our investigations reveal that employing either MLP-based~\cite{qi2017pointnet,qi2017pointnet++,wang2019dynamic} or MLP-Transformer~\cite{zhao2021point} based structures as the feature encode backbones consistently leads to improved performance when integrating the 2.5D map. Without special instructions, We use the DGCNN as the point cloud feature extractor for the other experiments.}
\begin{table}
\caption{Comparison between multi-modal methods using different point cloud encoders. The model parameters and Top-1 accuracy is calculated for three testing sets. Models denoted with * utilize the pixel-to-point fusion, whereas the remaining models adopt global fusion with the add operator. PT is the abbrev Point Transformer.}
\begin{center}
\small
\setlength{\tabcolsep}{1mm}
\begin{tabular}{lcccc}
\hline
Model & Params &  Hudson River & Wall Street & Union Square \\
\hline
Pointnet  & 283456 & 63.60 & 57.36 & 75.22 \\
Pointnet++ & 1335104  & 64.32 & 57.02 & 76.08 \\
PT & 7462464  & 64.02 & 58.56 & 77.26 \\
DGCNN & 1144192 & 64.08 & 56.26 & 76.54 \\
Pointnet* & 2907456  & 65.38 & 57.76 &  78.72 \\
\textbf{DGCNN*} & \textbf{3768648} & \textbf{67.70} & \textbf{60.66} & \textbf{82.96} \\
\hline
\end{tabular}
\end{center}
\label{table:backbone}
\end{table}

\vspace{2mm}
\noindent
\textbf{Embedding size}
{In this study, we investigate how the size of the embedding affects the single-image based localization. Table~\ref{table:dimension} shows that using PCA to reduce the feature dimension from 4096 to 128 only slightly lowers the performance. However, gradually reducing the feature dimensionality results in a more noticeable decline in performance. In particular, replacing 128-D feature embeddings with 16-D embeddings leads to a significant drop in performance. This suggests that low-dimensional representations are not discriminative enough to enable localization with a single image.}

\begin{table}
\caption{Comparison between multi-modal methods using 16-D to 4096-D semantic features in the area of Hudson River, Wall Street and Union Square. Top-1 recall rate (\%) is calculated to evaluate the localization performance.}
\begin{center}
\small
\setlength{\tabcolsep}{3mm}
\begin{tabular}{lccc}
\hline
Dimension & Hudson River & Wall Street & Union Square \\
\hline
16  &47.54 & 42.98 & 63.68 \\
32  & 59.68 & 53.88 & 77.30 \\
64  & 65.46 & 58.80 & 81.84 \\
128 & 67.70 & 60.66 & 82.96 \\
\textbf{4096}  & \textbf{68.28} & \textbf{61.26} & \textbf{83.10} \\
\hline
\end{tabular}
\end{center}
\label{table:dimension}
\end{table}

\vspace{2mm}
\noindent
\textbf{Culling strategy}
To improve efficiency, we eliminate 50\% of the route candidates at each movement until at least 100 remain. The impact of candidate culling on localization performance is shown in Fig.~\ref{fig:localization}(a). There is nearly no performance degradation in all testing areas. The results indicate that the culling strategy is efficient while preserving good localization capability.
In addition, the high similarity between curves also presents that route discrimination occurs early and is maintained as routes grow, which leads to a faster and stable convergence. We adopt a 50\% culling approach for the route based localization task. 

\vspace{2mm}
\noindent
\textbf{Semantic category}
{In Fig.~\ref{fig:semantic}, the 2.5D map comprises 24 distinct semantic categories. Certain mainstream methods~\cite{armagan2017learning,hai2021bdloc} solely employ building information to generate the 2.5D map for fine localization tasks. In this work, we investigate the performance enhancement achieved by incorporating richer semantic information within the 2.5D map.
As depicted in Fig.~\ref{fig:agg_opt}(c), there is a performance degradation of 10.08\% and 2.38\% for the Wall Street and Union Square areas, respectively. By including points from other semantic categories, such as water bodies and residential roads, the Top-1 accuracy further increases.
The results affirm the significance of incorporating diverse semantic information within the 2.5D map to achieve superior localization outcomes across varied urban landscapes, particularly in more sparsely built areas like Wall Street as shown in Fig.~\ref{fig:dataset}(b).}

\vspace{2mm}
\noindent
\textbf{Semantic label}
The 2.5D map input encompasses both 3-D coordinates ($x, y, z$) and corresponding semantic labels for each point. In the preceding experiments, only the coordinate information was utilized. To explore the impact of including explicit semantic information in the input data, we project the original 24-D one-hot vector into a 3-D learnable semantic encoding using a fully connected layer. This feature was then concatenated with the 3-D point coordinates to form the input for the subsequent network layers.
As indicated in Table~\ref{table:feature}, incorporating semantic labels in the input yields a performance improvement, although not  significant. These results suggest that integrating semantic information may offer additional benefits to the model's performance, albeit in a modest manner.
\begin{table}
\caption{Comparison between multi-modal methods with and without incorporating semantic labels as input. The Top-1 accuracy gains are shown in bracket.}
\begin{center}
\small
\setlength{\tabcolsep}{1mm}
\begin{tabular}{lccc}
\hline
Semantic Label & Hudson River & Wall Street & Union Square \\
\hline
-  & 67.70 & 60.66 & 82.96 \\
\checkmark  & \textbf{68.72 (+1.02)} & \textbf{60.88 (+0.33)} & \textbf{83.34 (+0.38)} \\
\hline
\end{tabular}
\end{center}
\label{table:feature}
\end{table}

\vspace{-4mm}
\subsection{Complexity Analysis}
We analyzed the computational cost and complexity of various methods on an Nvidia 3090 GPU by evaluating their Top-1 accuracy, inference time, memory utilization, and model size. Our findings are given in Table~\ref{table:model_efficiency}. Our multi-modal methods outperform single-modal approaches, with larger success rates and smaller model sizes, but they require more inference time and memory usage. The explicit geometric transform based method displays exceptional efficiency and performance.

When it comes to multi-modal methods that use various fusion strategies, the pixel-to-point fusion approach is the best in terms of localization performance, model size, and memory usage. However, it takes longer to infer. We compared the performance and efficiency gaps by varying the number of points in the multi-modal method in Table~\ref{table:npt_efficiency}. Fewer points may result in lower localization performance while improving efficiency.

\begin{table}
\begin{center}
\small
\setlength{\tabcolsep}{1mm}
\caption{Complexity comparison on Union Square between single-modal methods and multi-modal methods. Pol represents polar transformation. Memory is the maximum GPU memory occupied by tensors in an inference loop (batch size of 1). Size is the model size. 3to2 represents model with point-to-pixel fusion, while 2to3 means pixel-to-point fusion.}
% \vspace{-3mm}
\begin{tabular}{lcccc}
\hline
Model & Top-1 (\%) & Time (ms) & Memory (MB) & Size (MB) \\
\hline
ES & 56.06 & 2.37 & 53.20 & 378.73 \\
SAFA-Pol & 76.38 & 2.66 & 33.20 & 131.40 \\
Ours-concat & 74.10 & 4.45 & 94.94 & 263.58 \\
Ours-add & 76.54 & 4.35 & 89.61 & 199.58 \\
Ours-3to2 & 79.64 & 5.01 & 89.34 & 171.21\\
Ours-2to3 & 82.96 & 5.62 & 88.81 & 164.93\\
\hline
\end{tabular}
\end{center}
% \vspace{-9mm}
\label{table:model_efficiency}
\end{table}

\begin{table}
\begin{center}
\small
\setlength{\tabcolsep}{2mm}
\caption{Complexity comparison on Union Square between multi-modal methods using different number of points.}
% \vspace{-3mm}
\begin{tabular}{lccc}
\hline
Number of points & Top-1 (\%) & Time (ms) & Memory (MB) \\
\hline
2048 & 83.66 & 14.69 & 301.65 \\
1024 & 82.96 & 5.62 & 88.81 \\
512 & 81.12 & 3.56 & 36.61 \\
256 & 73.58 & 2.95 & 36.10  \\
\hline
\end{tabular}
\end{center}
\vspace{-3mm}
\label{table:npt_efficiency}
\end{table}

\section{Conclusion}
In this paper, we proposed ground-to-2.5D map matching for image-based geolocalization. Unlike previous methods, which only used 2D maps as the georeferenced database, we extended the 2D maps to 2.5D maps, where the heights of structures can be used to support cross-view matching. A new multi-modal representation learning framework is proposed to learn location embeddings from 2D images and point clouds. We also constructed the first large-scale ground-to-2.5D map geolocalization dataset to facilitate future research. Extensive experiments demonstrate that our multi-modal embeddings achieve significantly higher localization accuracy in both single-image based localization and route based localization.

% \section*{Acknowledgments}
% This should be a simple paragraph before the References to thank those individuals and institutions who have supported your work on this article.

\bibliographystyle{IEEEtran}
\bibliography{paper}

\newpage

\vfill

\end{document}